\documentclass{llncs}

\usepackage{graphicx}
\usepackage{amsmath, amssymb}
\usepackage{mathrsfs}
\usepackage{multirow}
\usepackage{enumitem}
\usepackage{pifont}
\usepackage{caption}
\usepackage{wrapfig}
\usepackage[misc]{ifsym}
\usepackage[skip=2pt]{caption}
\usepackage{algorithm}
\usepackage{algorithmic}
\renewcommand{\algorithmiccomment}[1]{\bgroup\hfill\tiny\#~#1\egroup}
\graphicspath{ {figs/} }
\usepackage{mathtools}
\DeclarePairedDelimiter\ceil{\lceil}{\rceil}

\newsavebox\IBoxA \newsavebox\IBoxB \newlength\IHeight
\newcommand\TwoFig[6]{% Image1 Caption1 Label1 Image2 ...
  \sbox\IBoxA{\includegraphics[width=0.48\textwidth]{#1}}
  \sbox\IBoxB{\includegraphics[width=0.48\textwidth]{#4}}%
  \ifdim\ht\IBoxA>\ht\IBoxB
    \setlength\IHeight{\ht\IBoxB}\else\setlength\IHeight{\ht\IBoxA}\fi%
  \begin{figure}[!htb]
  \minipage[t]{0.48\textwidth}\centering
  \includegraphics[height=\IHeight]{#1}
  \caption{#2}\label{#3}
  \endminipage\hfill
  \minipage[t]{0.48\textwidth}\centering
  \includegraphics[height=\IHeight]{#4}
  \caption{#5}\label{#6}
  \endminipage 
  \end{figure}%
}

\newcommand{\newpara}{\vspace{2mm}\noindent \textbf}

\usepackage{url}
\urldef{\mailsa}\path|{xingjunm@student., swijewickrem@, yun.zhou@,|
\urldef{\mailsb}\path|zhous@student., sjoleary@, baileyj@}unimelb.edu.au|
\newcommand{\keywords}[1]{\par\addvspace\baselineskip
\noindent\keywordname\enspace\ignorespaces#1}

\begin{document}
\setlength{\textfloatsep}{10pt}
\setlength{\intextsep}{10pt}
\mainmatter  % start of an individual contribution

\title{Providing Effective Real-time Feedback in Simulation-based Surgical Training}

% \titlerunning{A Random Forest Based Real-time Feedback Formulation Method}

\author{Xingjun Ma$^{\textrm{\Letter}}$
\and Sudanthi Wijewickrema \and Yun Zhou  \and\\ Shuo Zhou \and Stephen O'Leary \and James Bailey}
% index{Ma, Xingjun}
% index{Wijewickrema, Sudanthi}
% index{Zhou, Yun}
% index{Zhou, Shuo}
% index{O'Leary, Stephen}
% index{Bailey, James}

% \authorrunning{Xingjun Ma etc.}

\institute{The University of Melbourne, Melbourne, Australia\\
\mailsa\\
\mailsb\\
}

%\toctitle{Lecture Notes in Computer Science}
%\tocauthor{Authors' Instructions}
\maketitle

\begin{abstract}
Virtual reality simulation is becoming popular as a training platform in surgical education. However, one important aspect of simulation-based surgical training that has not received much attention is the provision of automated real-time performance feedback to support the learning process. Performance feedback is actionable advice that improves novice behaviour. In simulation, automated feedback is typically extracted from prediction models trained using data mining techniques. Existing techniques suffer from either low effectiveness or low efficiency resulting in their inability to be used in real-time. In this paper, we propose a random forest based method that finds a balance between effectiveness and efficiency. Experimental results in a temporal bone surgery simulation shows that the proposed method is able to extract highly effective feedback at a high level of efficiency.

\end{abstract}

\section{Introduction}
With the introduction of virtual reality (VR) technologies, simulation-based surgical training has become a powerful platform for surgical training \cite{rhienmora2011intelligent,wijewickremadesign,masimulation}. In VR simulation, trainees can practice on multiple surgical cases in a risk-free, immersive, and interactive learning environment. However, it still requires the supervision of human experts to ensure that feedback is provided during training to facilitate the acquisition of proper surgical skills. To reduce the reliance on the availability of human experts, the provision of feedback should be automated.

One of the simplest ways of providing real-time feedback in VR simulation is the ``follow-me" approach \cite{rhienmora2011intelligent}. It provides a `ghost' drill recorded by an expert to lead the trainee through the surgery. However, trainees who are unfamiliar with the procedure may struggle to follow the pace of the ghost expert. Other works utilised data mining techniques to generate feedback that can change adaptively in response to trainee's performance. For example, a pattern mining algorithm was applied to discover expert and novice skill patterns to support feedback provision in a surgical simulation \cite{zhou2013pattern}. However, experts and novices often share a considerable amount of similar behaviour which makes it difficult to identify significant patterns. This effect can be reduced through the use of random forest (RF) based methods such as  Split Voting (SV) \cite{zhou2013constructive}. SV first trains a RF prediction model that can distinguish expert skill from novice skill, then uses the prediction model to formulate feedback. Although this method is efficient, the effectiveness of formulated feedback in improving novice skill is low. In fact, extracting feedback from a RF is a NP-hard problem. However, it can be solved by high-performance ILP solvers when transformed to an integer linear programming (ILP) problem \cite{cui2015optimal}. This approach is highly effective in improving novice skills as it searches for the global optimal solution but the searching process is computationally expensive. Recent research shows that neural networks can also be used to generate feedback for simulation-based training \cite{ma2017extracting}.

% Using RF to extract actionable knowledge (or feedback) has been observed in other domains as well, such as customer relationship management \cite{yang2003postprocessing}.

There are three challenges in providing real-time feedback in a simulation environment: 1) feedback should be effective so it can improve novice skill to expert skill, 2) feedback should be provided in real-time (within 1 second) when novice skill is detected, as delayed feedback may result in confusion or cause undesirable consequences \cite{zhou2013constructive}, and 3) feedback should be simple (ideally based on one aspect of performance) so that the cognitive load is manageable \cite{sweller1988cognitive}.

The ideal feedback formulation method would be highly effective, yet efficient enough to be used in real-time. Highly effective methods exist but they typically lack sufficient efficiency and vice versa. Thus, the key in real-time feedback formulation is to find the right balance between the two, which to our knowledge has not been adequately addressed in the literature. To overcome this, we make the following contributions: 1) discuss the RF-based feedback formulation problem from a computational geometric point of view, 2) propose a novel method to formulate feedback using a RF, and 3) demonstrate that it has near-optimal effectiveness, is highly efficient, and scalable.

\section{Simulation Platform and Problem Definition}

The simulation used here is a \textit{Temporal Bone Surgery (TBS) Simulator} (see Fig. \ref{fig:tbs}). It consists of a 3D temporal (ear) bone model running on a computer and a haptic device providing tactile resistance to simulate drilling. The surgery we focus on is cortical mastoidectomy, which involves drilling parts of the temporal bone. In TBS, surgical skill is defined using characteristics of a \textit{stroke} (e.g. length, speed, acceleration, duration, straightness, force). A stroke is a continuous drilling motion in the same general direction that results in material removal \cite{zhou2015automated}. Feedback is the action that can improve a novice stroke to an expert stroke.   

%\vspace{-0.6cm}
\TwoFig{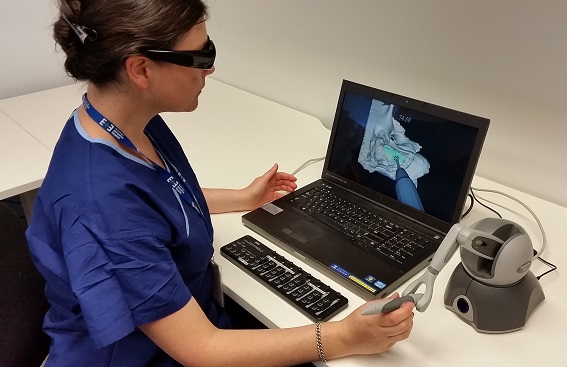}    {A temporal bone surgery simulator}    {fig:tbs}
      {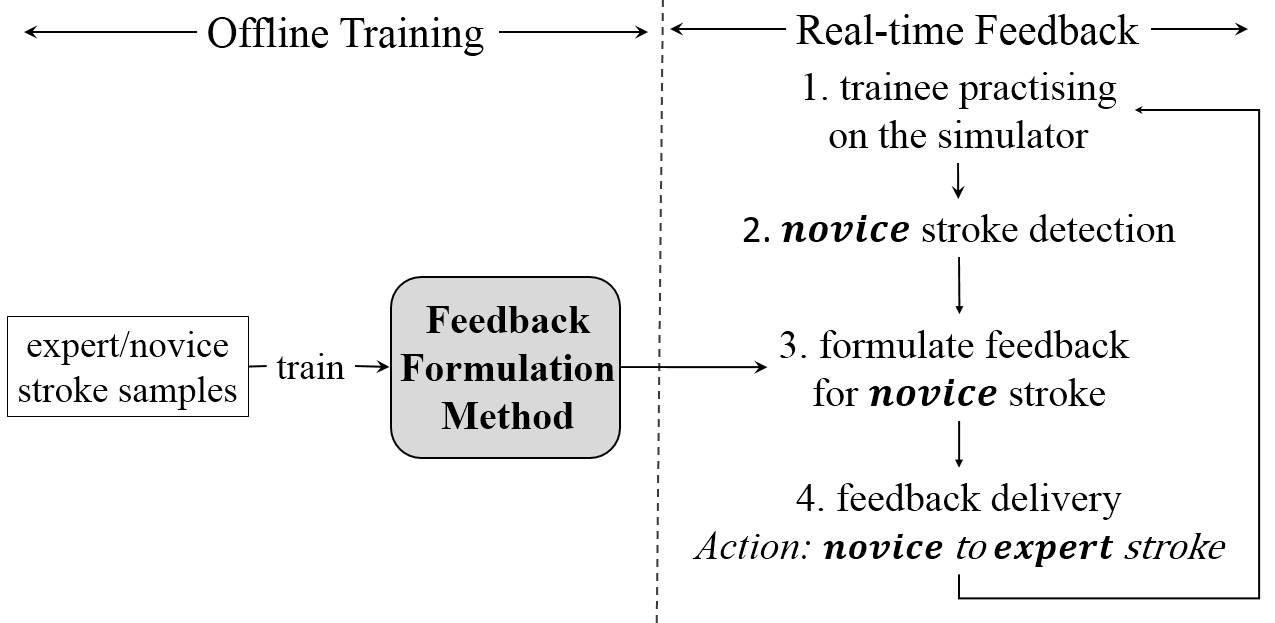}    {Real-time feedback formulation}    {fig:fbsystem}
%\vspace{-0.6cm}

An overview of the feedback formulation process is shown in Fig. \ref{fig:fbsystem}. The formulation method is trained offline over labelled (expert/novice) strokes. This is used in real-time to provide feedback that improves novice behaviour. The feedback formulation problem is to find the \textit{best change in stroke characteristics} that changes a novice stroke to an expert stroke. Let $\{X,Y\}$ be the dataset with $n$ strokes (or instances) and $d$ features (or stroke metrics). A stroke is defined by a feature vector $\mathbf{x}=(x_1,...,x_d)$ and is associated with a class label $y\in\{0,1\}$ (1: expert, 0: novice). The feedback formulation problem can then be defined as:

\begin{problem}
Given a prediction model $F(\mathbf{x})$ learnt over $\{X,Y\}$ and a novice stroke $\mathbf{x}$, the feedback formulation problem is to find the optimal action $A: \mathbf{x} \rightarrow \mathbf{x}_f$ that changes $\mathbf{x}$ to a stroke $\mathbf{x}_f$ under limited cost such that $\mathbf{x}_f$ has the highest probability of being in the expert class:
\label{prob:maximisation}
    \begin{equation*}
        \begin{aligned}
        & \underset{A:\; \mathbf{x} \rightarrow \mathbf{x}_f}{\text{argmax}}
        & & F(\mathbf{x}_f) \;\;
        & \text{subject to}
        & & \lVert \mathbf{x}-\mathbf{x}_f \rVert_0 \leq 1,
        \end{aligned}
    \end{equation*}
\end{problem}

\noindent where, $F(\mathbf{x}_f) \in [0,1]$ indicates the probability of $\mathbf{x}_f$ being in the expert class. Feedback $A:\mathbf{x} \rightarrow \mathbf{x}_f$ involves at most one feature change (increase/decrease). For example, $A:(force=\mathbf{0.2}, duration=0.3) \rightarrow (force=\mathbf{0.5},duration=0.3)$ is ``increase $force$ to 0.5". $\lVert \mathbf{x}-\mathbf{x}_f \rVert_0 \leq 1$ is the cost constraint that only allows a change in a single feature, to minimise cognitive load \cite{sweller1988cognitive}.

\section{Random Forest based Feedback Formulation}

We propose the use of RF as the prediction model $F(\mathbf{x})$. In contrast to existing methods \cite{zhou2013constructive,cui2015optimal,yang2007extracting}, we analyse the RF-based feedback formulation problem from a geometric point of view and introduce an approximation method to solve it. 

% As outlined in Algorithm \ref{algorithm:DA}, the DA method operates in three steps: 1) RF discretization, 2) hyper-rectangle pruning, and 3) discrete approximation. Step 1 discretizes the hyper-rectangles that define a RF into a finite number of discrete points so that it can be represented using a small number of points. Step 2 is to reduce the search space by pruning hyper-rectangles so as to solve the problem more efficiently. Step 3 uniformly selects a small number of points from the pruned hyper-rectangles as representatives and searches for the optimal solution within these representatives.

\begin{figure}[!htb]
    \centering
    \begin{minipage}{0.48\textwidth}
        \centering
        \includegraphics[scale=0.3]{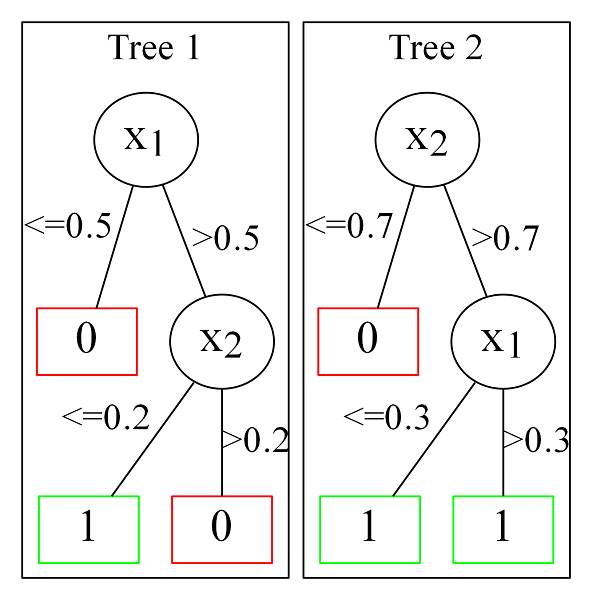}
        \caption{An example of a random forest. Colored view is recommended.}
        \label{fig:example_rf}
    \end{minipage}%
    \hfill
    \begin{minipage}{0.48\textwidth}
        \centering
        \includegraphics[scale=0.35]{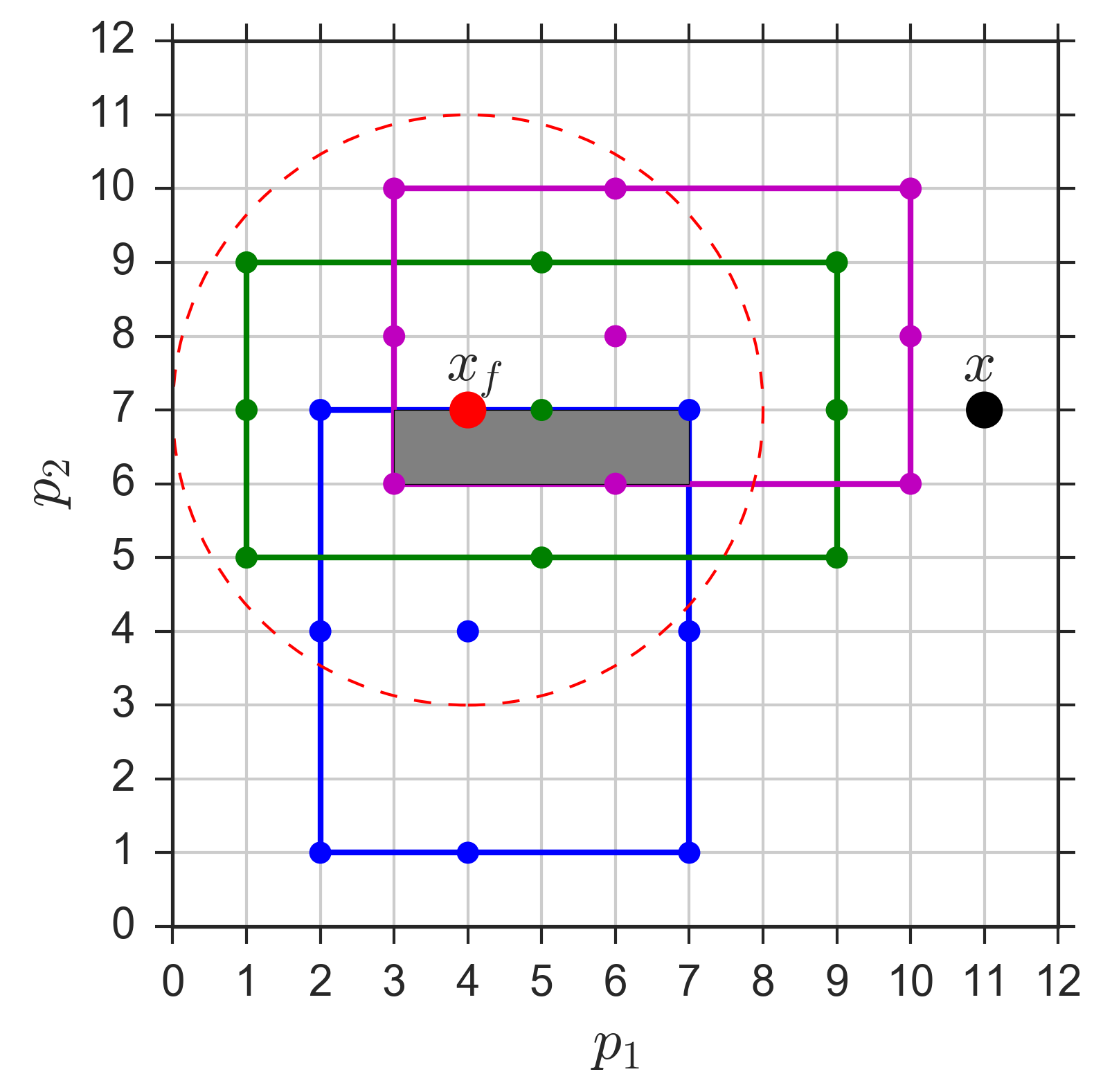}
        \caption{A 2D approximation example. Colored view is recommended.}
        \label{fig:three_da}
    \end{minipage}
\end{figure}

\newpara{Geometric View:} A RF prediction model is an ensemble of decision trees with each tree trained over a random subset of labelled strokes and features (see Fig. \ref{fig:example_rf}). The leaf node of the tree can be seen as a hyper-rectangle defined by the decision path from the root to the node. As in the example, the expert node (green rectangle with label 1) in Tree 1 can be represented by rectangle $\{x_1>0.5, x_2 \leq 0.2\}$. Thus, a RF can be seen as a union of hyper-rectangles overlapping in the data space \cite{breiman2000some}. Similar to a leaf node, a hyper-rectangle has an associated class label indicating the expertise level of strokes within it. Let $\mathcal{R}^{1}$/$\mathcal{R}^{0}$ denote the hyper-rectangle with a expert/novice class label respectively.
% ($R^{0} \in \mathcal{R}^{0}$, $R^{1} \in \mathcal{R}^{1}$, and $\mathcal{R}^{0} \cup \mathcal{R}^{1} = \mathcal{R}$).

A RF divides the data space into expert and novice subspaces. Expert subspaces are small areas that are overlapped by more $\mathcal{R}^{1}$s than $\mathcal{R}^{0}$s. Thus, the most expert-like strokes can be found in the areas that are overlapped by the most $\mathcal{R}^{1}$s. Then, feedback can be interpreted as moving a stroke from novice subspace to areas that are overlapped by the most $\mathcal{R}^{1}$s. However, calculating all possible intersections between $\mathcal{R}^{1}$s is NP-hard. To overcome this, we propose an approximation method as follows that uniformly takes a few points from $\mathcal{R}^{1}$ to represent them and finds a solution based on those representatives.

% \begin{algorithm}[H]
%  \caption{discrete approximation}
%  \label{algorithm:DA}
%  \begin{algorithmic}[1]
%  \renewcommand{\algorithmicrequire}{\textbf{Input:}}
%  \renewcommand{\algorithmicensure}{\textbf{Output:}}
%  \REQUIRE $\mathbf{x}$: novice stroke, $RF$: a pre-trained random forest prediction model
%   \STATE Extract expert hyper-rectangles $\mathcal{R}^{1}$s from $RF$
%   \STATE Discretize $\mathbf{x}$ and $\mathcal{R}^{1}$s to integer form \COMMENT{\textbf{Step 1: Random Forest Discretization}}
%   \STATE Remove or prune $\mathcal{R}^{1} \rightarrow \mathcal{R}^{*}$ for speed up \COMMENT{\textbf{Step 2: Hyper-rectangle Pruning}}
%   \FOR {each $\mathcal{R}^{*}$}
%       \STATE Extract representative points and add to set $\mathcal{P}$ \COMMENT{\textbf{Step 3: Discrete Approximation}}
%     %   \STATE Add extracted points into set $\mathcal{P}$
%   \ENDFOR
%   \STATE Search $\mathcal{P}$ for the optimal point ($\mathbf{x}_f$) that $\mathbf{x}$ can be moved to  %\COMMENT{\textbf{3} Discrete Approximation}
%  \ENSURE  feedback: $A: \mathbf{x}\rightarrow\mathbf{x}_f$
%  \end{algorithmic} 
%  \end{algorithm}

\vspace{-2mm}
\vspace{-2mm}

% \TwoFig{example_rf.png}    {An example of a random forest}    {fig:example_rf}
% {rectangle_three.png}{An approximation example} {fig:three_da}

\newpara{Random Forest Discretization:}
This step discretizes the expert hyper-rectangles of a RF to a finite number of integer-represented points. This transformation allows the uniform selection of a few points from an expert hyper-rectangle as its representatives. In a RF, a feature value is automatically segmented into multiple partitions by the split nodes. Suppose the number of partitions of the $d$ features are $m_1,...,m_d$ respectively. We define an integer variable $p_{i} \in [1,m_i]$ for feature $x_i$ to represent the indices of $x_i$'s partitions. For example, $p_{i}=j$ represents the $j^{th}$ partition of the $i^{th}$ feature. Thus, a stroke $\mathbf{x}=(x_1,x_2,...,x_d)$ can be written in integer form as $\mathbf{x}=(p_{1},p_{2},...,p_{d})$. As expert hyper-rectangle $\mathcal{R}^{1}$ is defined by the partition values, it can also be transformed to integer form as $\mathcal{R}^{1}=\{ l_1 < p_1 \leq r_1,..., l_d < p_d \leq r_d\}$, where $(l_i, r_i]$ defines the integer form of the value range in dimension $x_i$.

% \vspace{-2mm}
% \begin{equation}
% \label{eq:int_inst}
% \mathbf{x}=(p_{1},p_{2},...,p_{d})
% \end{equation}
% \vspace{-2mm}

% \vspace{-2mm}
% \begin{equation}
% \label{eq:int_hyper}
% R^{1}=\{l_1 < p_1 \leq r_1,l_2 < p_2 \leq r_2,...,l_d < p_d \leq r_d\}
% \end{equation}
% \vspace{-2mm}

In Fig. \ref{fig:example_rf}, suppose $x_1, x_2 \in [0,1]$. Then, based on where it splits, $x_1$ has partitions: $[0,0.3]^1$, $(0.3,0.5]^2$ and $(0.5,1]^3$
while $x_2$ has partitions: $[0,0.2]^1$, $(0.2,0.7]^2$ and $(0.7,1]^3$. Thus, a stroke $\mathbf{x}=(x_1=0.4, x_2=0.8)$ can be transformed into integer form as $\mathbf{x}=(p_{1}=2, p_{2}=3)$ with $p_{1}$ and $p_{2}$ denoting the partition number for their respective features. The expert rectangle $\mathcal{R}^{1}=\{x_1>0.5, x_2 \leq 0.2\}$ can be transformed into $\mathcal{R}^{1}=\{ 2 < p_{1} \leq 3, 0 < p_{2} \leq 1\}$. As the RF grows each tree on a random subset of features, the number of features that defines a hyper-rectangle may be less than the total number of features.

With this transformation, a stroke is discretized to an integer point and an expert hyper-rectangle is discretized to a finite number of points. For example, $\mathcal{R}^{1}=\{1 < p_1 \leq 3,2 < p_2 \leq 3\}$ is equivalent to $\mathcal{R}^{1}=\{ p_1 \in \{2, 3\}, p_2 \in \{3\}\}$ denoting only two points in $\mathcal{R}^{1}$: $(p_1=2, p_2=3)$ and $(p_1=3, p_2=3)$. As RF can deal with both numerical and categorical data, this transformation can be applied to arbitrary datasets. For simplicity, in the rest of this paper, ``stroke'' and ``hyper-rectangle'' denote the discretized integer forms.

\newpara{Hyper-rectangle Pruning:}
This process reduces the search space by 1) removing redundant expert hyper-rectangles, and 2) removing redundant points within the remaining hyper-rectangles so as to increase computational efficiency. We denote the set of hyper-rectangles that remain after this process by $\mathcal{R}^{*}$. The RF feedback formulation problem can now be solved approximately by finding a new stroke that is in the densest overlapping area of $\mathcal{R}^{*}$s.

\vspace{4pt}
\noindent \textit{Removing redundant expert hyper-rectangles:} For an expert hyper-rectangle to be used to formulate feedback for a novice stroke $\mathbf{x}$, it should contain at least one possible solution, i.e., one point that satisfies the cost constraint: $\exists \mathbf{x}_f \in \mathcal{R}^{1}$ such that $\lVert \mathbf{x}-\mathbf{x}_f \rVert_0 \leq 1$. When formulating feedback for $\mathbf{x}$, expert hyper-rectangles that cannot provide any possible solutions can be removed. For example, given $\mathbf{x}=(p_1=2, p_2=3)$, the expert hyper-rectangle $\mathcal{R}^{1}=\{0 < p_1 \leq 1, 4 < p_2 \leq 5\}$ does not contain any feasible solutions as $\mathbf{x}$ has to change two features ($p_1$ and $p_2$) to be moved into this hyper-rectangle.

\vspace{4pt}
\noindent \textit{Pruning the remaining expert hyper-rectangles:} Further, not all points in the remaining hyper-rectangle are feasible solutions. Consider the same novice stroke as above and $\mathcal{R}^{1}=\{1 < p_1 \leq 3, 4 < p_2 \leq 5\}$. We can change $\mathbf{x}$'s $p_2$ to $5$ to move it into $\mathcal{R}^{1}$. However, there are other possible changes that can also move $\mathbf{x}$ into $\mathcal{R}^{1}$, but require more than one feature change (e.g. changing $p_1$ to $3$ and $p_2$ to $5$ at the same time). Such changes violate the cost constraint, and as such can be pruned from the solution space by fixing the value of $p_i$ in $\mathcal{R}^{1}$ to its value in the novice stroke $\mathbf{x}$, i.e., $\mathcal{R}^{1} \rightarrow \mathcal{R}^{*}=\{\mathbf{p_1 = 2}, 4 < p_2 \leq 5\}$.

\newpara{Discrete Approximation:}
Finding points that are in the densest overlapping area of $\mathcal{R}^{*}$s can be done by iterating through all points in each $\mathcal{R}^{*}$. However, this approach is again computationally expensive. We avoid this through a discrete approximation (DA) method that uniformly selects a small number of points as representatives for $\mathcal{R}^{*}$, and takes the center of the densest area of these representatives as an approximation of the optimal solution.

We introduce a parameter $\alpha \in (0,1]$ to indicate the proportion of representative values selected from each dimension of a hyper-rectangle $\mathcal{R}^{*}=\{l_1 < p_1 \leq r_1,l_2 < p_2 \leq r_2,...,l_d < p_d \leq r_d\}$. The number of selected values from the $i^{th}$ dimension can be calculated by:
\vspace{-2mm}
\begin{equation}
\label{eq:num_dim}
n_i = \begin{cases} 
r_i - l_i &\mbox{if } \ r_i - l_i \le 2 \\
\ceil*{\alpha(r_i-l_i)} + 2 & \mbox{otherwise }  \end{cases}
\end{equation}
\vspace{-2mm}

For a dimension $p_i$ that contains only 1 or 2 values, we directly use these values to represent $p_i$. Otherwise, we divide $[l_i+1,r_i]$ into $n_i+1$ equal segments and take the values in the division positions with $l_i+1$ and $r_i$ as representative values of $p_i$. After extracting these values out of each dimension, we derive $\prod_{i=1}^{d} n_i$ number of points by taking all the possible combinations.

To find the center of the densest area with respect to the representative points, we consider an area defined by a hyper-sphere with radius $\gamma$. Then, based on a computed Euclidean distance matrix between all representative points, the point which has the most neighbours with a distance less than or equal to $\gamma$ is selected. The selected point is transformed from integer form back to its original form by taking the corresponding original value in the partition position. The feedback is then constructed based on the original form. 
%The center can be found by a computed Euclidean distance matrix between all representative points that is calculated to find the center is the most time-consuming operation in the DA method with a time complexity of $O((\prod_{i=1}^{D} n_i)^2)$.

Fig. \ref{fig:three_da} illustrates an example where there are 3 expert rectangles (blue, green and purple) and 2 features: $x_1$ and $x_2$ ( $p_1, p_2$ are the integer forms of $x_1, x_2$ respectively). When taking 3 values in each dimension to represent a rectangle, we get 9 representative points per rectangle. Thus, with a radius $\gamma=4$, we can find the red center point $\mathbf{x}_f$ of the densest area (the red dashed circle) of representative points. As shown in the figure, $\mathbf{x}_f$ is also located in the best expert space (the grey area overlapped by all three expert rectangles) that $\mathbf{x}$ can be changed to. If the original form of $\mathbf{x}$ is $(x_1=\mathbf{0.9},x_2=0.2)$ and $\mathbf{x}_f$ is $(x_1=\mathbf{0.4},x_2=0.2)$, then the feedback for $\mathbf{x}$ is ``decrease $x_1$ to \textbf{0.4}". In a nutshell, this method utilises 9 points to approximate each rectangle and the red circle with a tunable radius to approximate the grey overlapping area. The center of the circle is used as an approximation of the solutions in the grey area.

%\begin{algorithm}[H]
% \caption{Discrete approximation algorithm}
% \label{algorithm:DA}
% \begin{algorithmic}[1]
% \renewcommand{\algorithmicrequire}{\textbf{Input:}}
% \renewcommand{\algorithmicensure}{\textbf{Output:}}
% \REQUIRE $\mathbf{x}^{n}$: novice stroke, $R^{(1)}$: expert hyper-rectangles, $C$: cost limit, $\alpha$, $\gamma$
% \ENSURE  $\mathbf{x}$
%  \STATE $\mathscr{P}$: point set
%  \FOR {$R_{i}^{(1)}$ in $R^{(1)}$}
%      \IF {$\mathscr{C}(\mathbf{x}, R_{i}^{(1)}) \leq C$}
%          \STATE $\mathscr{V}$ = value sets for each dimension
%          \STATE $R^{f}$ = prune($R_{i}^{(1)}$)
%          \FOR {$l_i < p_i \leq r_i$ in $R^{f}$}
%          \STATE $n_i = \ceil*{\alpha(l_i-r_i)}$
%          \STATE $\mathscr{V}_i$ = {$n_i$ equal division values of $(l_i,r_i]$}
%          \ENDFOR
%          \STATE add all possible points in $\mathscr{V}$ to $\mathscr{P}$
%      \ENDIF
%  \ENDFOR  
%  \STATE M = pairwiseManhattanDistance($\mathscr{P}$)
%  \STATE $\mathbf{x}$ = findCenter(M, $\gamma$)
% \RETURN $\mathbf{x}$
% \end{algorithmic}
% \end{algorithm}

\section{Experimental Evaluation} \label{sec:experiment}

We compared the performance of the proposed DA method with 5 other methods:  

\vspace{-2mm}
\begin{enumerate}
\item \textbf{Split Voting (SV):} is the state-of-the-art real-time feedback formulation method for RF introduced in \cite{zhou2013constructive}.
\item \textbf{Integer Linear Programming (ILP):} solves the RF feedback formulation problem by transforming it to an integer linear programming problem \cite{cui2015optimal}.
\item \textbf{Random Selection (Rand-Rand):} randomly picks a feature from a novice stroke and selects a random partition of that feature as the feedback.
\item \textbf{Random Iteration (Rand-Iter):} randomly selects a feature and iteratively selects the best partition to change it to as the feedback \cite{cui2015optimal}.
\item \textbf{Iterative Selection (Iter-Iter):} iteratively tests all partitions of each feature while keeping the other features fixed. The overall best combination of feature and partition is selected as the feedback \cite{cui2015optimal}.
\end{enumerate}
\vspace{-2mm}

The parameters of the DA method were tuned on the training data using a grid search. The parameter values that showed the best results under the real-time efficiency requirement of 1 second ($\alpha=0.5$ and $\gamma=2$) were chosen. The methods were evaluated based on the following 3 measures:

\vspace{-2mm}
\begin{enumerate}
\item \textbf{Success Rate (SR):} $SR = \frac{|\{\ \mathbf{x}_f| \, F(\mathbf{x}_f)>0.5\}|}{|\{\mathbf{x} \}|}$ is the percentage of novice strokes that are successfully changed to expert strokes. Higher values of SR denote better performance.
%\begin{equation}
%\label{eq:successrate}
%SR = \frac{|\{\ \mathbf{x}_f| \, F(\mathbf{x}_f)>0.5\}|}{|\{\mathbf{x} \}|}
%\end{equation}
\item \textbf{Effectiveness (EFF):} $EFF = F(\mathbf{x}_f)$ is the value of the objective function defined in Problem \ref{prob:maximisation}, i.e., the probability of the target stroke being an expert stroke. It measures how effective the feedback is when applied to change the novice stroke. Higher values signify better effectiveness.
%\begin{equation}
%\label{eq:effectiveness}
%EFF = F_{y=1}(\mathbf{x}_f)
%\end{equation}
\item \textbf{Time-cost (TC):} Time (seconds) spent to formulate one feedback, lower is better. The lower the time-cost, the higher the efficiency.
\end{enumerate}
\vspace{-2mm}

\setcounter{footnote}{0} Experiments were carried out on a typical PC with a 2.40GHz CPU. The ILP solver used for the ILP method was CPLEX\footnote{https://www-01.ibm.com/software/commerce/optimization/cplex-optimizer} as recommended by the authors \cite{cui2015optimal}. Our dataset consisted of 60K strokes (28K expert strokes and 32K novice strokes) performed by 7 experts and 12 novices. All methods were evaluated on a RF with 100 trees and a maximum depth of 5. A 12-fold leave-one-novice-out cross-validation was used to obtain an unbiased measure. In each fold, we used all strokes performed by one novice as the test set and trained a RF prediction model on the remaining strokes. All methods were then applied to formulate feedback for strokes in the test set using the trained RF. This design simulates the real-world scenario of an unknown novice using the simulator.

As shown in Table \ref{table:effect_compare}, in terms of success rate and effectiveness, the proposed method (DA) is comparable to Iter-Iter and ILP that find optimal solutions and outperforms the other methods by a large margin. However, Iter-Iter and ILP take more than 10 seconds to formulate one feedback, and therefore fail to meet the real-time efficiency requirement of 1 second. Overall, DA achieves near-optimal success rate and effectiveness, and is highly efficient, indicating that it finds the ideal balance for real-time use in TBS simulation. 

\newpara{Parameter and Scalability Analysis:}
As shown in Fig. \ref{fig:alpha_DA}, the time-cost of the proposed method DA increases consistently with the increase of the parameter $\alpha$. However, the effectiveness is stabilized after $\alpha=0.5$. As  $\alpha$ controls the number of representative points, this indicates that using more points for the approximation is not necessary and the effectiveness-efficiency balance can be achieved by the use of a smaller $\alpha$. We also tested the scalability of all methods with regard to the number of trees in the RF. As shown in Fig. \ref{fig:scalability_trees}, the time-cost of ILP increases dramatically as the number of trees increases and it takes more than 4 minutes to formulate one feedback using a RF with 1,000 trees. Iter-Iter is also seen to have a considerable increase in time-cost with the increase of the number of trees. However, the other methods, including the proposed method (DA), are more stable and remain highly efficient for large scale RFs.

% \vspace{-2mm}
\begin{table}[t]
\renewcommand{\arraystretch}{1.5}
\caption{Performance (mean$\pm$std) comparison between DA and other methods. The best results are highlighted in bold.}
\label{table:effect_compare}
\centering
\begin{tabular}{c||c|c|c|c|c|c}
\hline
 & Rand-Rand & Iter-Iter & Rand-Iter & ILP & SV & DA \\ \hline \hline
success rate & 0.21$\pm$0.04 & \textbf{0.89$\pm$0.00} & 0.36$\pm$0.05 & \textbf{0.89$\pm$0.00} & 0.60$\pm$0.05 & \textbf{0.89$\pm$0.00} \\ \hline
effectiveness  & 0.18$\pm$0.23 & \textbf{0.87$\pm$0.06} & 0.40$\pm$0.30 & \textbf{0.87$\pm$0.06} & 0.65$\pm$0.33 & 0.84$\pm$0.08 \\ \hline
 time-cost (s) & \textbf{0.00$\pm$0.00} & 12.17$\pm$0.14 & 0.36$\pm$0.05 & 32.07$\pm$2.57 & 0.02$\pm$0.00 & 0.26$\pm$0.15 \\ \hline
\end{tabular}
\end{table}
% \vspace{-2mm}

% More specifically, $\alpha=0.5$ means taking $\frac{1}{2}$ of the values out of each dimension as representatives. Since we have 6 dimensions (stroke is defined by 6 features, e.g., d=6), overall the representatives only count for $\frac{1}{2^6}$ of all points in the target areas. Therefore, DA should be 32 times faster than Iter-Iter or ILP (global search among all points), and this is proved by the results in Table \ref{table:effect_compare}.

% \vspace{-2mm}
\TwoFig{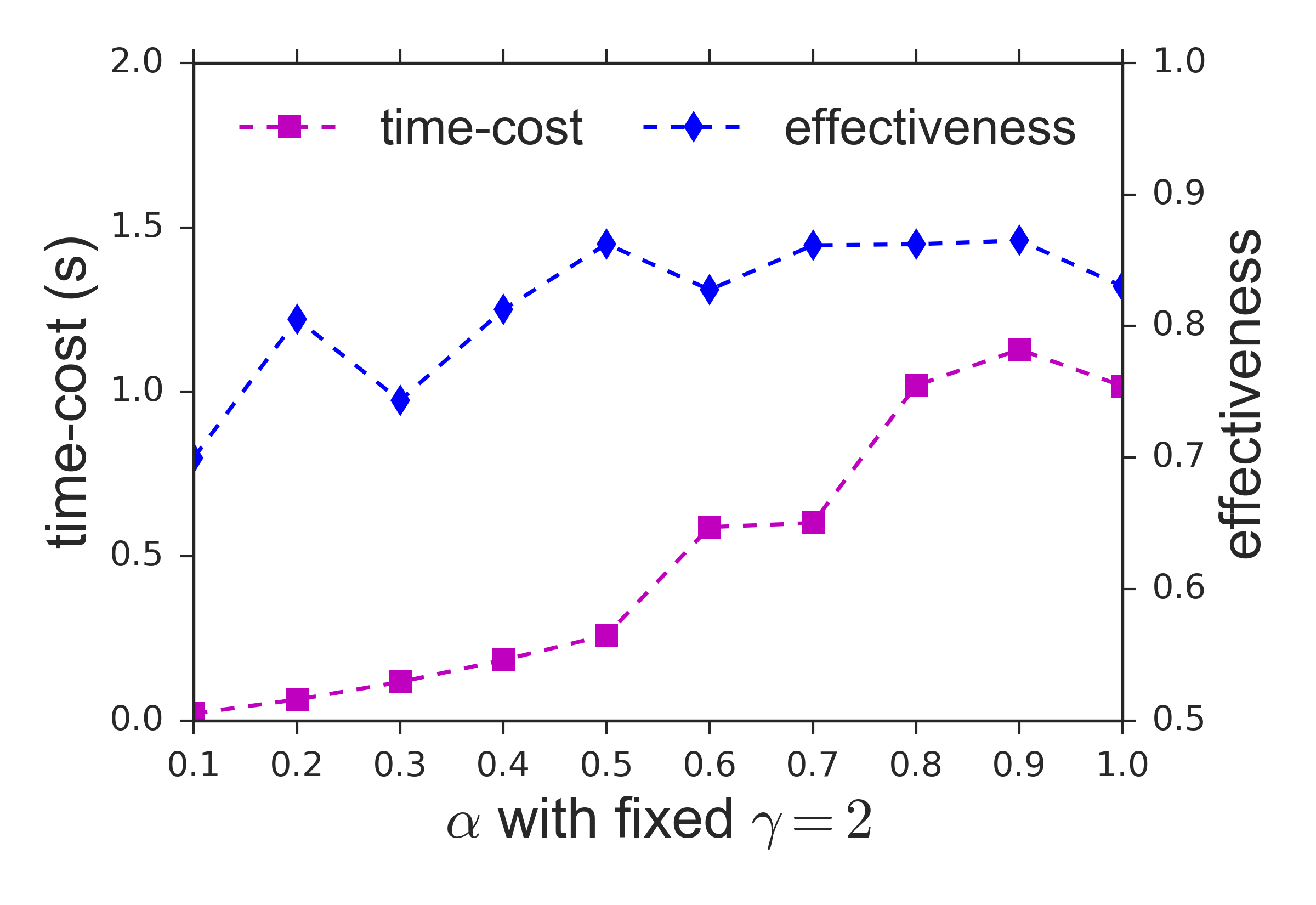}    {Performance of DA w.r.t. $\alpha$}    {fig:alpha_DA}
{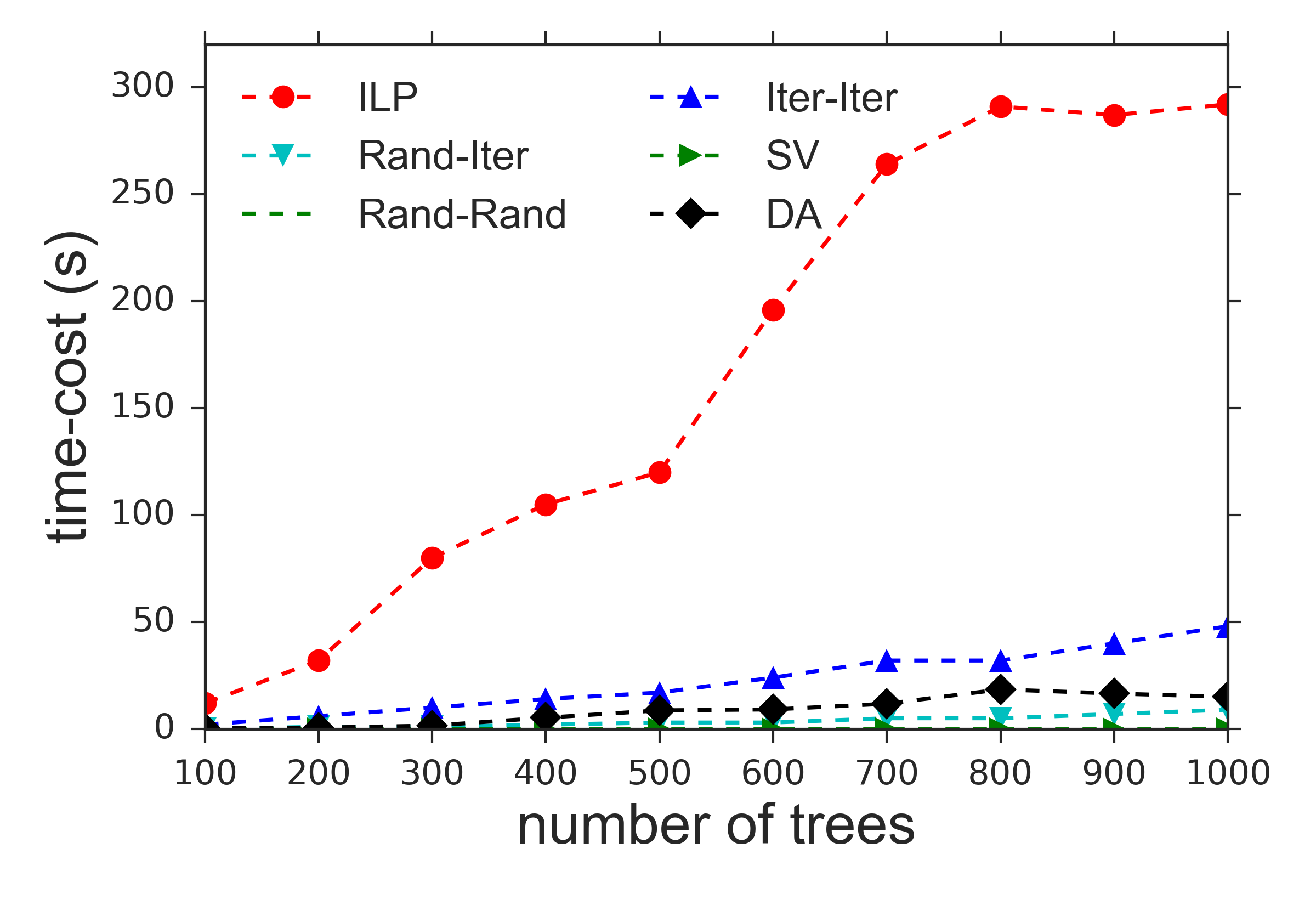}    {Scalability over number of trees}    {fig:scalability_trees}
% \vspace{-2mm}

% \vspace{-2mm}
% \begin{figure}[!ht]
% \centering
% \includegraphics[width=0.5\textwidth]{scalability.png}
% \caption{Scalability over number of trees.}
% \label{fig:scalability_trees}
% \end{figure}
% \vspace{-2mm}

\section{Conclusion}
In this paper, we discussed the problem of formulating feedback using a random forest based model for the specific application of virtual reality temporal bone surgery simulation. We discretized the hyper-rectangles of a random forest into integer form and proposed a novel method to formulate feedback using these hyper-rectangles. We also showed that the proposed method outperformed the state-of-the-art methods in terms of success rate and effectiveness while remaining highly efficient. Moreover, it is consistently efficient for large-scale random forests. The proposed method can be generalized to other simulation training platforms where real-time feedback is of importance.

\section*{Acknowledgements}
This research has received support from the Office of Naval Research Global.

\bibliographystyle{splncs03unsrt}
\bibliography{paper1148}

\end{document}